\pdfoutput=1
\documentclass{article}
\usepackage{spconf,amsmath,graphicx,hyperref}
\usepackage{placeins} 
\usepackage{subfigure}

\usepackage{spconf}
\usepackage{amsmath,amssymb,amsfonts,graphicx}
\usepackage{placeins}
\usepackage[utf8]{inputenc}
\usepackage[all]{hypcap}

\hypersetup{
    colorlinks=true,
    linkcolor=blue,
    citecolor=blue,
    urlcolor=blue
}

\title{ADAPTIVE GUIDANCE SEMANTICALLY ENHANCED VIA MULTIMODAL LLM FOR EDGE-CLOUD OBJECT DETECTION}
\name{Yunqing Hu\textsuperscript{\rm 1,3}, Zheming Yang\textsuperscript{\rm 1}, Chang Zhao\textsuperscript{\rm 1,3},Wen Ji\textsuperscript{\rm 1,2}}
\address{ \textsuperscript{\rm 1}Institute of Computing Technology, Chinese Academy of Sciences, Beijing, China \\ \textsuperscript{\rm 2}Institute of AI for Industries, Nanjing, China  \\ \textsuperscript{\rm 3}University of Chinese Academy of Sciences, Beijing, China   }

\usepackage{color}

\begin{document}
\ninept
\maketitle
\begin{abstract}
Traditional object detection methods face performance degradation challenges in complex scenarios such as low-light conditions and heavy occlusions due to a lack of high-level semantic understanding. To address this, this paper proposes an adaptive guidance-based semantic enhancement edge-cloud collaborative object detection method leveraging Multimodal Large Language Models (MLLM), achieving an effective balance between accuracy and efficiency. Specifically, the method first employs instruction fine-tuning to enable the MLLM to generate structured scene descriptions. It then designs an adaptive mapping mechanism that dynamically converts semantic information into parameter adjustment signals for edge detectors, achieving real-time semantic enhancement. Within an edge-cloud collaborative inference framework, the system automatically selects between invoking cloud-based semantic guidance or directly outputting edge detection results based on confidence scores. Experiments demonstrate that the proposed method effectively enhances detection accuracy and efficiency in complex scenes. Specifically, it can reduce latency by over 79\% and computational cost by 70\% in low-light and highly occluded scenes while maintaining accuracy.

\end{abstract}


\section{Introduction}
\label{sec:intro}
Object detection is a fundamental computer vision task with applications in autonomous driving, security, and medical imaging \cite{zhao2019object}\cite{ragab2024comprehensive}\cite{chen2021deep}. Deep learning models such as YOLO series \cite{redmon2016you}\cite{jiang2022review}, SSD \cite{liu2016ssd}, Faster R-CNN \cite{Girshick_2015_ICCV}, and Mask R-CNN \cite{he2017mask} have significantly improved detection speed and accuracy, enabling real-time inference. However, these vision-feature-driven approaches struggle in complex scenarios like low light, heavy occlusion, and dense crowds, leading to lower recall and more false positives \cite{ahmed2021survey}\cite{rahman2018zero}\cite{pourpanah2022review}due to their reliance on fixed labels and pixel-level features without high-level semantic understanding \cite{xiao2020noise}\cite{dhillon2020convolutional}.

Multimodal Large Language Model (MLLM) demonstrates potential for open-word detection and contextual reasoning by integrating visual and linguistic inference, thereby supplementing the semantic limitations of traditional detectors \cite{rouhi2024enhancing}. Representative works, including ContextDET \cite{zang2025contextual}, DetGPT \cite{pi2023detgpt}, and VOLTRON \cite{wase2023object}, have advanced semantic augmentation and cross-modal fusion. However, directly applying MLLM still faces challenges such as low regional accuracy, high computational overhead, and unstructured output \cite{wang2024qwen2}\cite{chen2020deep}, making it difficult to fully replace lightweight detectors such as YOLOv12 \cite{tian2025yolov12} and RT-DETR \cite{zhao2024detrs}. To address this problem, academia has proposed detection frameworks combining LLM and light models. Examples include YOLO-World \cite{cheng2024yolo} enabling open-vocabulary detection via CLIP, Ferret \cite{you2023ferret} enhancing candidate region semantics with language models, and LLMDet \cite{fu2025llmdet} jointly training detectors and language models. However, these approaches still have limitations: fusion mechanisms are often statically designed, unable to dynamically adapt to real-time scenarios; moreover, the semantic information generated by LLM is insufficiently structured, hindering efficient utilization at the edge.

To address these challenges, we propose an adaptive prompt-based semantic enhancement framework for edge-cloud collaborative object detection using MLLM. The framework integrates cloud-level semantic reasoning with lightweight edge detection to dynamically balance accuracy and real-time performance. The MLLM is instruction-tuned to produce structured JSON outputs, overcoming free-text variability and hallucinations. A lightweight mapping module then converts these semantics into core parameters for real-time optimization of edge detectors. During inference, the system adaptively switches between cloud-enhanced and edge-only detection based on confidence scores, as illustrated in Fig. \ref{fig:fig1}. The main contributions are as follows:

\begin{figure*}
    \centering
    \includegraphics[width=1\linewidth]{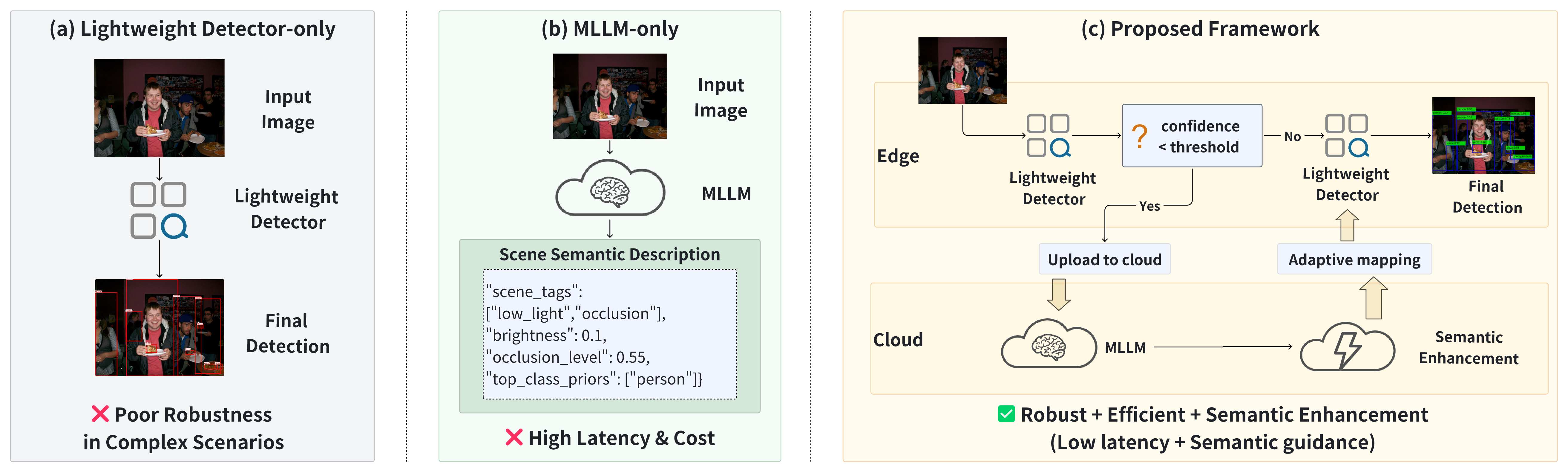}
    \caption{Comparison of different framework. (a) Lightweight Detector-only, offering low latency but weak semantics; (b) MLLM-only, with strong semantics but high latency and cost; (c) Our edge-cloud collaborative architecture, combining efficient edge detection with cloud semantic guidance for optimal accuracy and efficiency.}
    \label{fig:fig1}
\end{figure*}

\begin{itemize}
\item We introduce a novel instruction MLLM fine-tuning paradigm and design an adaptive mapping mechanism that dynamically converts semantic guidance into parameter adjustment signals for lightweight edge detectors. 
    
\item We develop an edge-cloud collaborative inference framework equipped with a confidence-based decision strategy. The system automatically switches between invoking cloud-level semantic enhancement and directly outputting edge results, effectively balancing latency and accuracy. 
    
\item  Experiments demonstrate that this method can reduce latency by over 79\% and computational cost by 70\% in low-light and highly occluded scenes with minimal compromise to accuracy.

\end{itemize}

\section{Method}
\label{sec:format}
\begin{figure}
    \centering
    \includegraphics[width=1\linewidth]{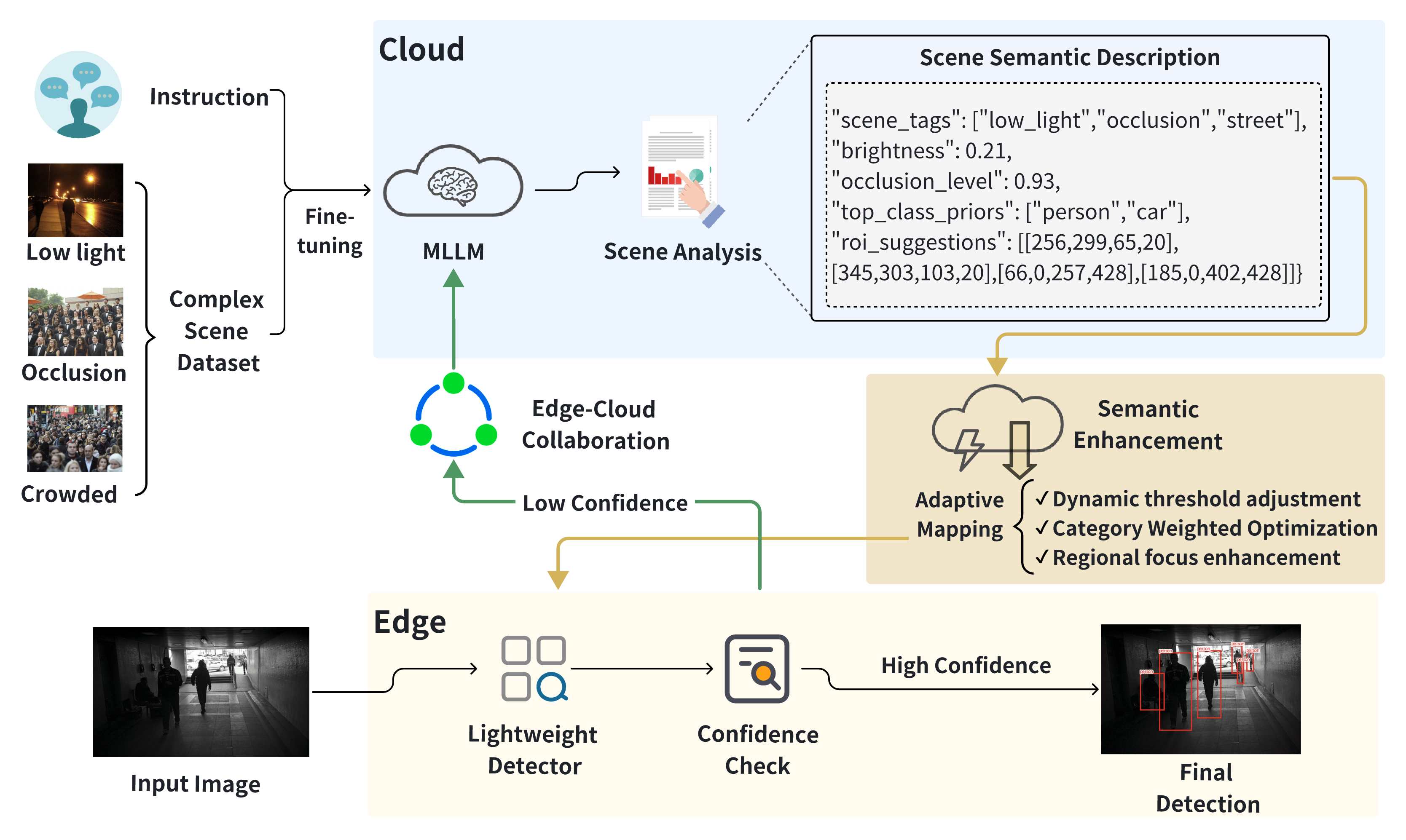}
    \caption{Overall architecture of the proposed adaptive guidance semantically enhanced object detection framework.}
    \label{fig:fig2}
\end{figure}

This paper proposes an adaptive guidance-enhanced semantic edge-cloud collaborative object detection framework. Its core architecture is shown in Fig. \ref{fig:fig2}, comprises three key modules: the Instruction-Tuned Semantic MLLM, the Adaptive Semantic-to-Parameter Mapping module, and the Edge-Cloud Collaborative Routing mechanism. This framework deeply integrates edge detection and cloud semantic inference, achieving a balance between accuracy and efficiency in complex scenarios.

\subsection{Instruction-Tuned Semantic MLLM}
MLLM outputs are typically unstructured free-form text, lacking formal expression and optimization for local region perception in complex scenarios, making them difficult to directly utilize for precise control of downstream detectors. To address this, we propose a structured instruction fine-tuning strategy enabling MLLM to simultaneously output bounding boxes and scene semantic descriptions adhering to a strict JSON format, thereby significantly enhancing the detector's adaptability in complex environments.

To ensure efficient scalability, we adopt Low-Rank Adaptation (LoRA) as a lightweight fine-tuning method. The core principle of LoRA involves training only low-rank matrices while keeping pre-trained weights frozen, thereby reducing parameter overhead and improving adaptability. Its update form can be expressed as:
\begin{equation}
W^{\prime}=W+\Delta W, \Delta W=A B^{T}, A \in \mathbb{R}^{d \times r}, B \in \mathbb{R}^{d \times r},
\end{equation}
where $W$ represents the original pre-trained weights, and $\bigtriangleup W$ denotes the low-rank update with rank $r\ll d$. This design preserves the general semantic capabilities of large models while enabling adaptation to specific detection tasks at a low cost. By jointly optimizing bounding box prediction and semantic consistency through the loss function, the model achieves structured semantic modeling for complex scenes while maintaining lightweight efficiency.

\subsection{Adaptive Semantic-to-Parameter Mapping}
Traditional lightweight detectors have fixed inference parameters that are unable to dynamically adjust to scene semantics, leading to performance degradation in complex real-world scenarios such as low-light conditions or high occlusion. This module aims to transform the structured semantic description $S$ into dynamic control signals for the detector, enabling adaptive optimization. The semantic description $S= \left \{ b,o,p,P,R \right \} $ comprises brightness $b\in \left [ 0,1 \right ] $, occlusion ratio $o\in \left [ 0,1 \right ] $, scene category prior distribution $P\left ( c \right ) $, and a set of recommended regions of interest (ROIs) $R$. Based on this information, we design three complementary mechanisms.

\subsubsection{Dynamic Threshold Adjustment}
Dynamic classification threshold adjustment modifies the baseline threshold $\tau _{0}$ based on brightness and occlusion information. MLLM introduces semantic variables $b$ and $o$. When low illumination or severe occlusion is detected, the classification threshold $\tau_c$ dynamically adjusts to reduce false negatives while balancing false positives. The adjusted threshold is calculated as:
\begin{equation}
\tau_{c}=\tau_{0}-\alpha_{1} \cdot(1-b)-\alpha_{2} \cdot o.
\end{equation}

\subsubsection{Category Weight Optimization}
Category weight optimization dynamically adjusts the weight $\omega_c$ for each category $c$. Where $\omega_0$ is the baseline weight, $I(\cdot)$ is an indicator function which activates when the estimated person count $p$ exceeds a threshold $p_{th}$, $o$ represents the occlusion level, and $P(c)$ denotes the semantic prior for category $c$. The hyperparameters $\beta_1$, $\beta_2$, and $\beta_3$ control the contributions of person density, occlusion, and semantic priors, respectively. The adaptive weight for each category is calculated as:
\begin{equation}
\omega_{c}=\omega_{0}+\beta_{1} \cdot I\left(p>p_{t h}\right)+\beta_{2} \cdot o+\beta_{3} \cdot P(c).
\end{equation}

\subsubsection{Region Focus Enhancement}
Region focus enhancement weights the scores of overlapping candidate boxes based on the ROI proposals obtained through semantic reasoning, thereby amplifying detection responses in critical regions. The region weighting function is:
\begin{equation}
G(x, y)=\left\{\begin{array}{l}
\gamma, \text { if }(x, y) \in R. \\
1, \text { otherwise. }
\end{array}\right.
\end{equation}

\subsection{Edge-Cloud Collaborative Routing}
To balance latency and accuracy, we propose a dynamic edge-cloud routing mechanism. A lightweight edge detector handles low-latency inference, while a fine-tuned cloud MLLM provides detection and semantic outputs as needed. The system selects local or cloud inference based on real-time confidence and scene metrics.

Defining the average confidence of edge model detection results as $\bar{C} $, the routing decision function is as follows:
\begin{equation}
f_{\text {route }}(I)=\left\{\begin{array}{l}
\text { Edge }- \text { only }, \bar{C} \geq \tau. \\
\text { Cloud }- \text { enhanced }, \text { otherwise. }
\end{array}\right.
\end{equation}
 
When edge detection confidence is high and scene complexity is low, the task is fully completed at the edge. When confidence is low or scene complexity is high, the task is uploaded to the cloud for semantic enhancement by MLLM, then returned to the edge for adaptive adjustment.

\section{Experiments}
\label{sec:experiments}
\subsection{Experimental Setup}
\textbf{Datasets and Models.}
To comprehensively evaluate the effectiveness of the proposed method, we conducted systematic experiments across multiple public datasets covering general scenes, low-light environments, and high-density occlusion scenarios. The experiments employed a hybrid dataset D = $<$COCO2017 $\cup $ ExDark $\cup $ CrowdHuman$>$   to ensure the model's generalization capability across varying levels of complexity. COCO 2017 \cite{COCO} provides a general object detection benchmark, ExDark \cite{loh2019getting} focuses on low-light and nighttime images, while CrowdHuman \cite{shao2018crowdhuman} contains numerous crowd occlusion scenarios. Each sample is organized as a triplet: $<i m g, I, O>$, where $img$ denotes the input image, $I$ represents the instruction text, and $O$ is the desired output comprising a set of bounding boxes $B=\left\{b_{i}\right\}_{i=1}^{N}$ and a structured semantic description set $S=\left\{s_{k}\right\}_{k=1}^{M}$  to ensure consistency between task inputs and outputs. We employed Qwen2-VL-7B as the multimodal large model foundation and YOLOv12s as the edge detector to compare performance across different configurations.

\textbf{Evaluation Metrics.}
The evaluation process is carried out from multiple dimensions. The compliance of the JSON output is measured by the completeness of the structure and the accuracy of the fields. For semantic accuracy, we calculated MSE for brightness estimation, F1 scores for scene labels, and MAE for person counting. Detection performance is primarily evaluated by mAP, recall, and F1 scores. System efficiency was evaluated through inference latency (ms), throughput (FPS), and computational consumption.

\textbf{Baseline Methods.}
In the main comparative experiment, we evaluated three system configurations: the edge-only approach, which processes images using YOLOv12s; the cloud-only approach, where all images are processed by Qwen2-VL-7B; and the edge-cloud collaborative optimization system proposed in this paper. Additionally, we conducted ablation experiments for each strategy in the adaptive semantic-to-parameter mapping mechanism to analyze the contribution of each optimization strategy.

\subsection{Experimental Results}

\subsubsection{Multimodal LLM Fine-tuning Results}

\begin{figure}[th]
	\centering 
    \subfigure[Semantic Accuracy]{
    \label{Fig.sub.3.1}
    \includegraphics[width=0.22\textwidth]{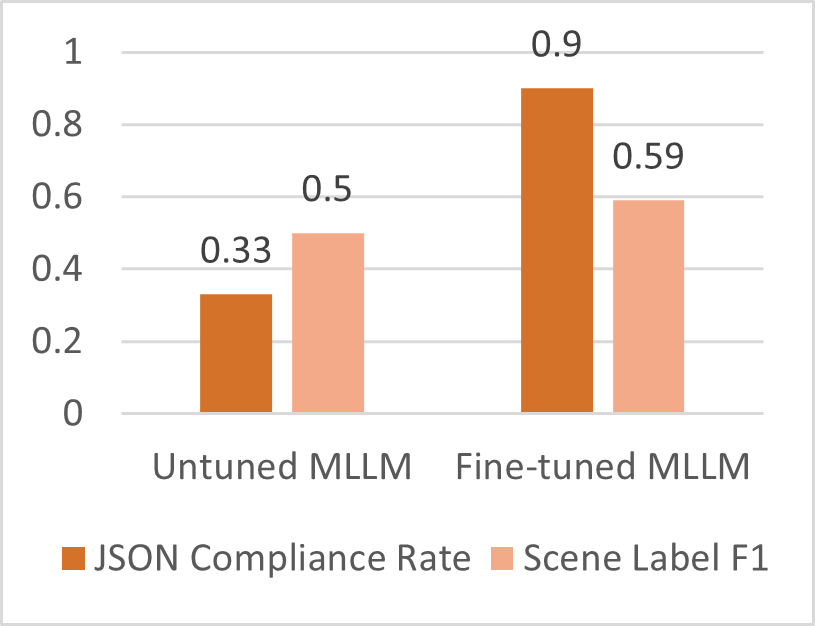}}
    \subfigure[Perceptual Quality]{
    \label{Fig.sub.3.2}
    \includegraphics[width=0.22\textwidth]{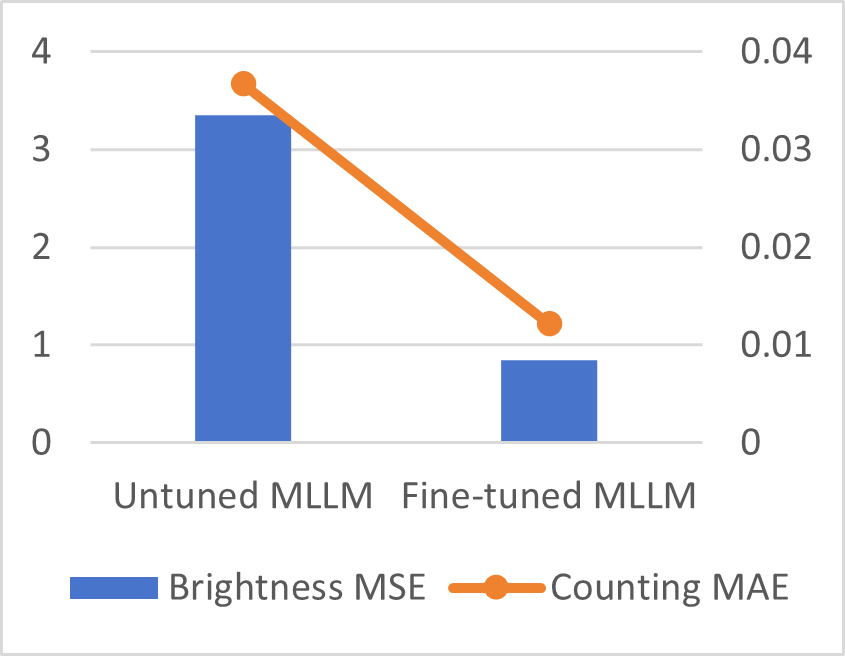}}
    
    \caption{Qwen2-VL-7B fine-tuning effects on (a) semantic accuracy and (b) perceptual quality.}
	\label{fig:fig3}
\end{figure}

Fig. \ref{fig:fig3} compares the performance of models before and after fine-tuning across multiple metrics for structured output and semantic understanding. For semantic accuracy, semantic compliance rate rises sharply from 0.33 to 0.90, and Scene Label F1 improves from 0.50 to 0.59, indicating stronger structured output and semantic understanding. For perceptual quality, brightness MSE drops from 0.0335 to 0.0085, and counting MAE decreases from 3.67 to 1.22, showing enhanced visual estimation and crowd counting accuracy.

\begin{figure}[th]
    \centering
    \includegraphics[width=1\linewidth]{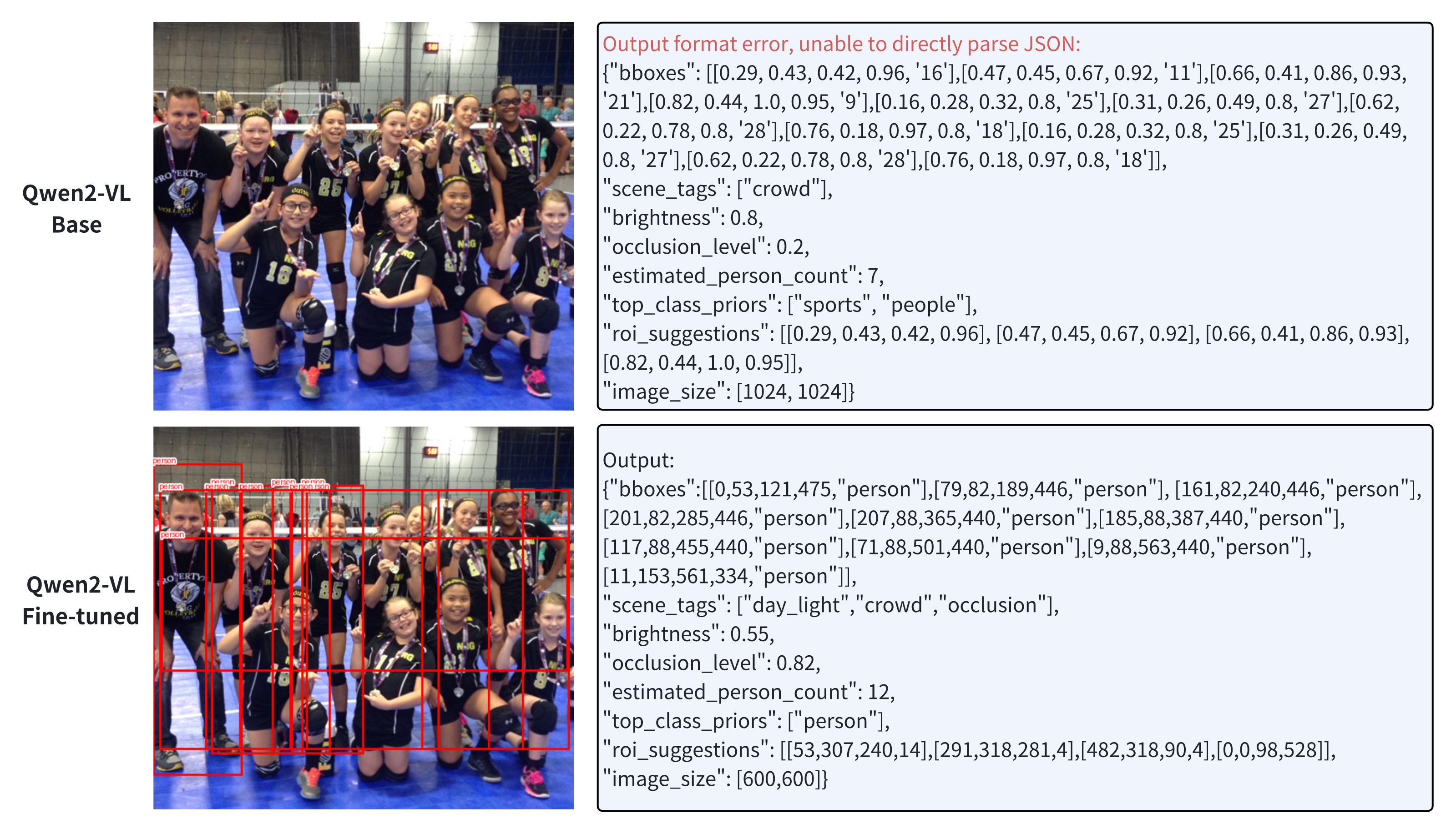}
    \caption{Output comparison before and after fine-tuning.}
    \label{fig:fig4}
\end{figure}

Output example from high-density scenes before and after fine-tuning further validates these results. As shown in Fig. \ref{fig:fig4}, the untuned model produces outputs with non-standard formats that are difficult to parse. In contrast, the fine-tuned model strictly adheres to JSON specifications while maintaining high programmability. It also generates more accurate and complete target localization and scene semantic information, highlighting the practical application potential of MLLM.

\subsubsection{Accuracy in Complex Scenarios}
To comprehensively evaluate the adaptability and robustness of the proposed method in complex real-world scenarios, we conducted systematic performance comparison experiments on a portion of the test sets of the ExDark dataset (low-light environments) and the CrowdHuman dataset (high-density occlusion scenarios). In terms of accuracy, the proposed method demonstrates outstanding detection performance across two complex scenarios, as shown in Table \ref{table1} and Table \ref{table2}. Compared to the edge-based YOLOv12s model, it achieves improvements of 5.7\% and 6.4\% on the ExDark and CrowdHuman datasets, respectively, significantly mitigating performance degradation caused by environmental interference. Compared to the cloud-based Qwen2-VL-7B model, the accuracy loss is minimal and negligible.


\begin{table}[!ht]
    \centering
    \caption{Detection performance on the ExDark dataset.}
    \begin{tabular}{|l|l|l|l|}
    \hline
        \textbf{Model} & \textbf{mAP@50} & \textbf{Recall} & \textbf{F1-Score} \\ \hline
        YOLOv12s & 0.775 & 0.718 & 0.76 \\ \hline
        Qwen2-VL-7B & 0.835 & 0.776 & 0.80 \\ \hline
        Ours & 0.832 & 0.767 & 0.79 \\ \hline
    \end{tabular}
\label{table1}
\end{table}
\begin{table}[h]
    \centering
    \caption{Detection performance on the CrowdHuman dataset.}
    \begin{tabular}{|l|l|l|l|}
    \hline
        \textbf{Model} & \textbf{mAP@50} & \textbf{Recall} & \textbf{F1-Score} \\ \hline
        YOLOv12s & 0.788 & 0.735 & 0.78 \\ \hline
        Qwen2-VL-7B & 0.843 & 0.788 & 0.81 \\ \hline
        Ours & 0.852 & 0.802 & 0.82 \\ \hline
    \end{tabular}
\label{table2}
\end{table}

\subsubsection{Real-time Performance in Complex Scenarios}
\begin{figure}[th]
	\centering 
    \subfigure[ExDark]{
    \label{Fig.sub.5.1}
    \includegraphics[width=0.22\textwidth]{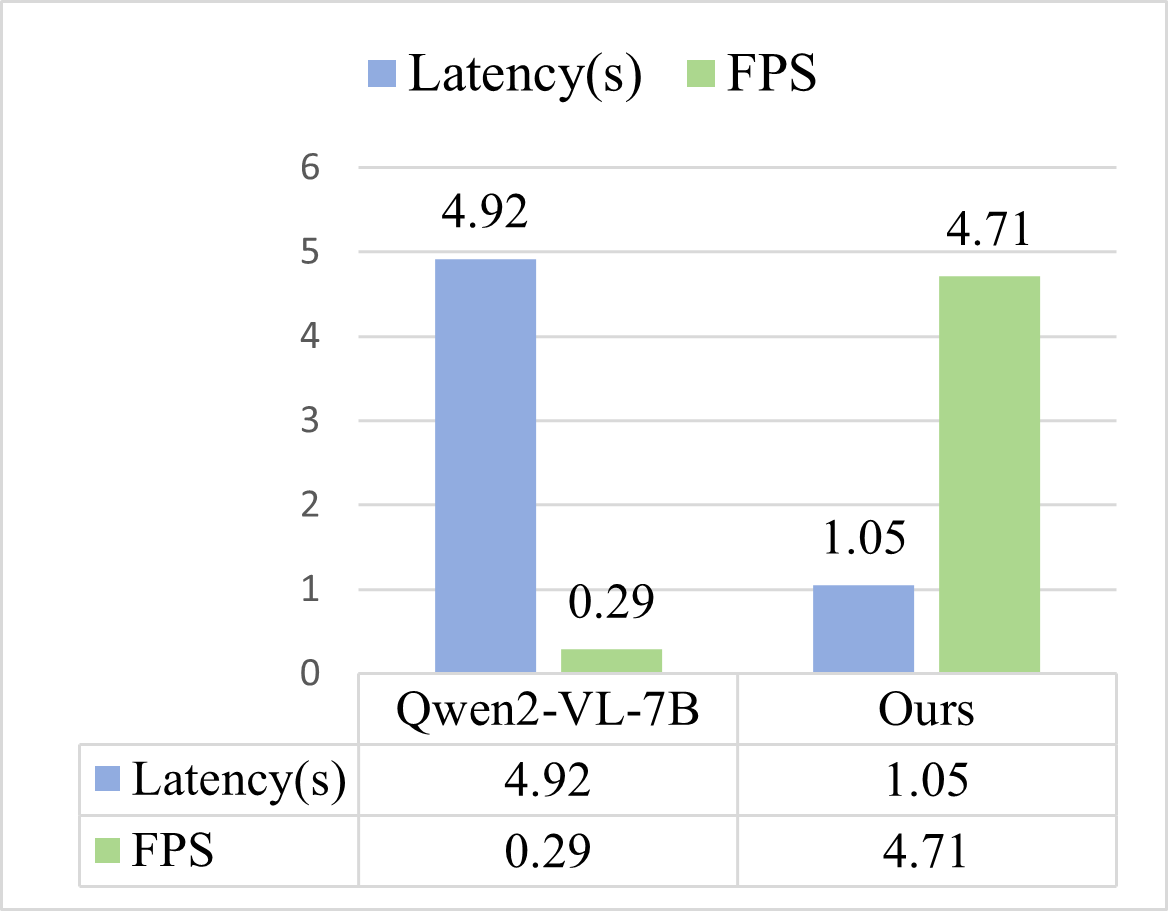}}
    \subfigure[CrowdHuman]{
    \label{Fig.sub.5.2}
    \includegraphics[width=0.22\textwidth]{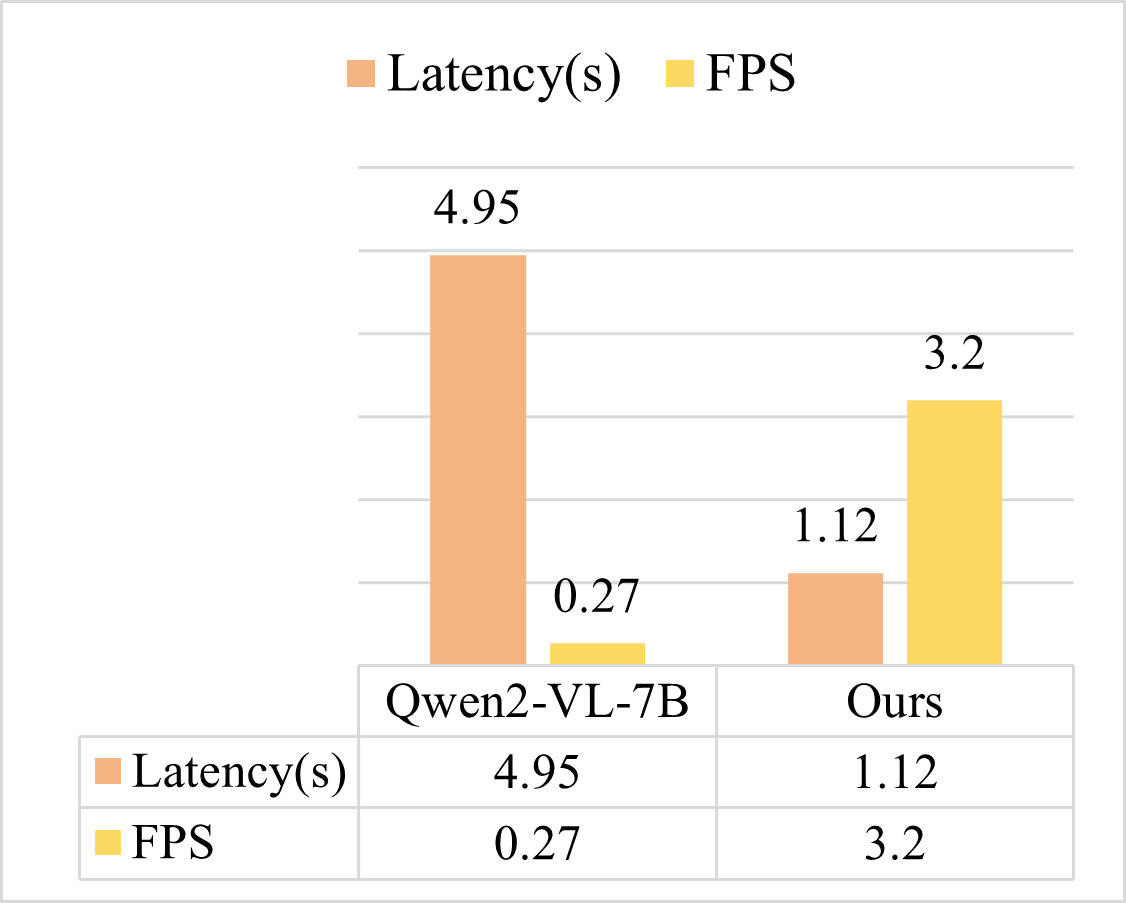}}
    
    \caption{Comparison of system real-time performance among different datasets. (a) shows results on the ExDark dataset, while (b) shows results on the CrowdHuman dataset.}
	\label{fig:fig5}
\end{figure}

In terms of real-time performance, our method demonstrates significant advantages. As shown in Fig. \ref{fig:fig5}, cloud-based inference incurs approximately 5 seconds of latency with an FPS below 0.3, rendering it incapable of real-time response. Our approach reduces latency to 1.05s and 1.12s while boosting FPS to 4.71x and 3.20x, achieving a 79\% latency reduction and enabling near-real-time inference in complex scenarios. The relatively consistent latency across different datasets demonstrates the robustness of the proposed method, suggesting that the adaptive edge-cloud collaboration effectively mitigates the computational bottleneck of cloud inference while maintaining semantic guidance benefits. Furthermore, the ability to preserve semantic guidance while offloading computations adaptively demonstrates that efficiency gains do not come at the expense of accuracy. 

\subsubsection{Resource Consumption in Complex Scenarios}
\begin{figure}[th]
	\centering 
    \subfigure[ExDark]{
    \label{Fig.sub.6.1}
    \includegraphics[width=0.22\textwidth]{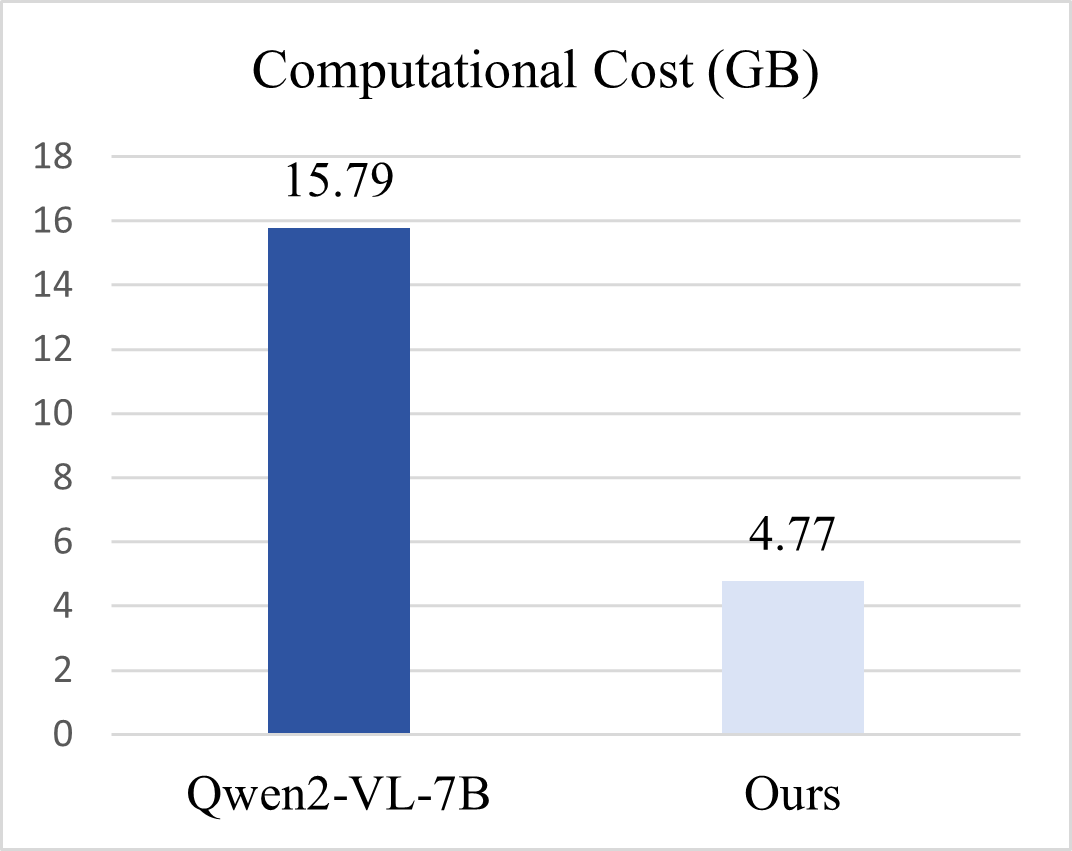}}
    \subfigure[CrowdHuman]{
    \label{Fig.sub.6.2}
    \includegraphics[width=0.22\textwidth]{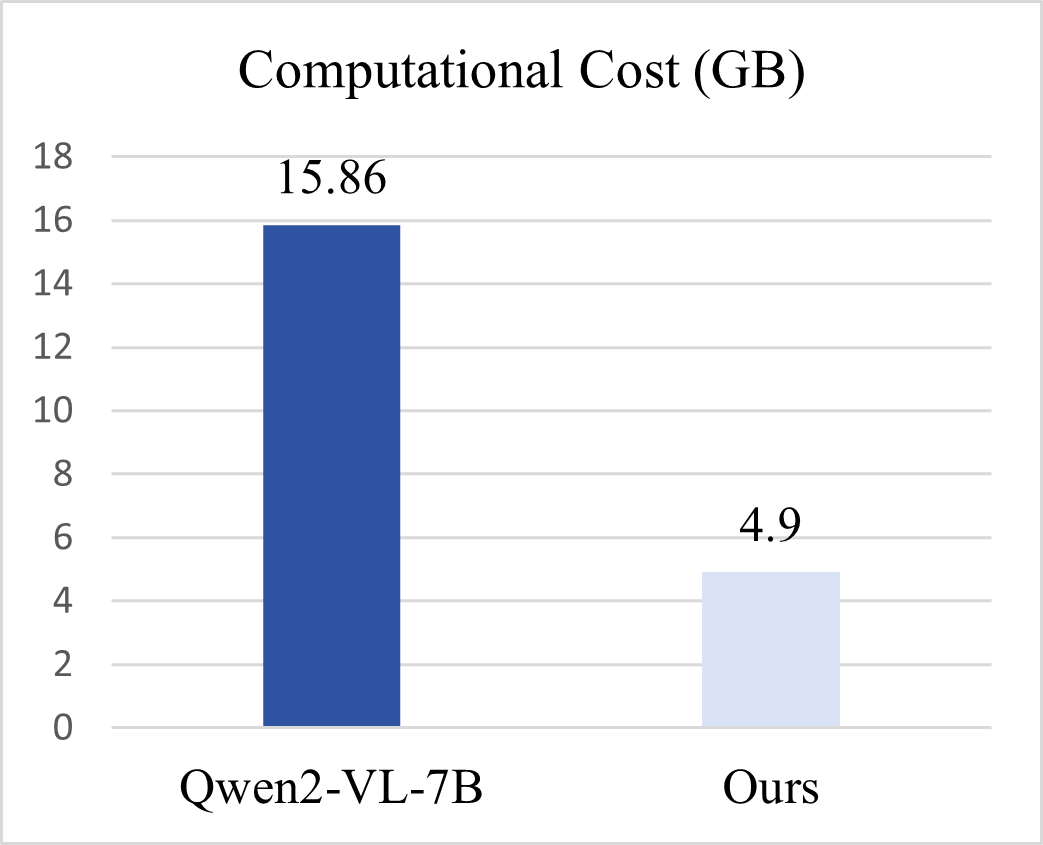}}
    
    \caption{Comparison of system resource consumption among different atasets. (a) shows results on the ExDark dataset, while (b) shows results on the CrowdHuman dataset.}
	\label{fig:fig6}
\end{figure}

Fig. \ref{fig:fig6} illustrates that our proposed method substantially reduces computational overhead by nearly 70\% compared with Qwen2-VL-7B. This dramatic reduction not only alleviates the hardware burden but also makes real-time edge deployment feasible under constrained resources. The results highlight the effectiveness of our adaptive guidance strategy in balancing workload distribution between edge and cloud, thereby avoiding the typical bottlenecks observed in conventional large-model deployments. Beyond efficiency gains, the comprehensive experimental results further demonstrate that our framework achieves a well-balanced tradeoff among accuracy, latency, and efficiency in complex scenarios. Together, these findings validate the robustness and practicality of the proposed method for edge-cloud object detection applications.



\subsubsection{Ablation Studies}
To evaluate the independent and combined contributions of each strategy within the adaptive semantic-to-parameter mapping module, we conducted ablation experiments on a mixed validation set consisting of ExDark and CrowdHuman. YOLOv12s served as the baseline, achieving an mAP50 of 0.587, a recall of 0.804, and an F1 score of 0.592. Results demonstrate that each strategy provides effective improvements. The dynamic threshold adjustment strategy increased mAP50 to 0.619, the category weight optimization strategy improved recall to 0.826, and the region focus enhancement strategy raised the F1-score to 0.625. Combining strategies yielded greater gains, with the joint application of threshold adjustment and class optimization strategies increasing F1-score by 6.33\%, demonstrating the complementary effects of semantic optimization and spatial attention. When all strategies were integrated, mAP50 rose by 5.69\% and F1 by 7.23\%, confirming the effectiveness of collaboration among multiple strategies for enhanced semantic understanding.

\section{CONCLUSION}
This paper proposes an adaptive MLLM-based semantic enhancement framework for edge-cloud collaborative object detection, effectively integrating the semantic understanding capabilities of MLLM with the efficient inference capabilities of lightweight detectors. Through structured semantic outputs, adaptive semantic-to-parameter mapping, and dynamic edge-cloud routing, the framework balances accuracy and efficiency in complex scenarios. Experiments demonstrate that this approach achieves 5.7\% and 6.4\% higher mAP than edge detection in low-light and high-occlusion scenarios, respectively, while significantly reducing latency and resource consumption compared to cloud-based solutions. Ablation studies validate the effectiveness of each mapping strategy. This research provides a practical solution for high-precision object detection in complex environments.


\bibliographystyle{IEEEbib}
\bibliography{strings,refs}

\end{document}